\documentclass[11pt]{article}

\usepackage{epsfig}
\usepackage{amssymb}
\usepackage{amsmath}
\usepackage{pstricks}
\usepackage{graphicx,epstopdf}

\usepackage{pdflscape}

\usepackage{enumerate}
\usepackage{tikz}
\usepackage{ulem}
\usepackage{caption}
\usepackage{epigraph}

\usepackage{algorithm}
\usepackage{algpseudocode}
\makeatletter
\def\BState{\State\hskip-\ALG@thistlm}
\makeatother

\definecolor{lightblue}{rgb}{.80,1,1}
\definecolor{myblue}{rgb}{.5,.5,.8}
\definecolor{dblue}{rgb}{0,.5,1}
\definecolor{lightpurple}{rgb}{1,.80,1}
\definecolor{lightyellow}{rgb}{1,1,.70}
\definecolor{webgreen}{rgb}{0,.5,0}
\definecolor{lightgrey}{rgb}{.8,.8,.8}

    \newtheorem{theorem}{Theorem}[section]
    \newtheorem{lemma}[theorem]{Lemma}

    \newtheorem{corollary}[theorem]{Corollary}

    \newtheorem{remark}[theorem]{Remark}
    
    \newtheorem{definition}[theorem]{Definition}
    
   \newcommand{\beq}{\begin{equation}}
\newcommand{\eeq}{\end{equation}}
\newcommand{\real}{{\mathbb R}}

\begin{document}

\title{\bf The Phylogenetic LASSO and the Microbiome\thanks{This research was funded in part by grants from:  CIHR-NSERC Collaborative Health Research Projects 413548-2012; NSERC RGPIN 46204-11; NSF grants  DBI-1262351
and DMS-1418007; NSF DMS-1127914 to SAMSI; and, PSI Foundation Health Research Grant 2013.}}
\author{
Stephen T Rush\thanks{
PhD candidate in the Department of Mathematics and Statistics, University of Guelph, Guelph, Ontario N1G 2W1 Canada.}\ \ 
Christine H Lee\thanks{MD and Professor, Department of Pathology and Molecular Medicine, McMaster University, St Joseph's Healthcare, 50 Charlton Ave E, 424 Luke Wing, Hamilton, Ontario L8N 4A6 Canada.}\ \ 
Washington Mio\thanks{
PhD and Professor, Department of Mathematics, Florida State University, Tallahassee,
FL 32306-4510 USA.}\ \ 
Peter T Kim\thanks{PhD, Professor and Corresponding Author, Department of Mathematics and Statistics, University of Guelph, Guelph, Ontario N1G 2W1 Canada, and, Department of Pathology and Molecular Medicine, McMaster University, St Joseph's Healthcare, 50 Charlton Ave E, 424 Luke Wing, Hamilton, Ontario L8N 4A6 Canada.} }
\date{}

\maketitle

\begin{abstract}
\noindent Scientific investigations that incorporate next generation sequencing involve analyses of high-dimensional data where the need to organize, collate and interpret the outcomes are pressingly important. Currently, data can be collected at the microbiome level leading to the possibility of personalized medicine whereby treatments can be tailored at this scale. In this paper, we lay down a statistical framework for this type of analysis with a view toward synthesis of products tailored to individual patients. Although the paper applies the technique to data for a particular infectious disease, the methodology is sufficiently rich to be expanded to other problems in medicine, especially those in which coincident `-omics' covariates and clinical responses are simultaneously captured.\end{abstract}

\vspace{.25 in}


\newpage

\baselineskip = 20pt plus 3pt minus 3pt

\section{Introduction}

The incorporation of  the microbiome as a covariate in response to biological or clinical outcomes
is becoming medically and scientifically important especially 
in understanding etiology and treatment of specific diseases.  In addition, this aids in improved understanding of functional
pathways such as the gut-brain axis, see for example \cite{shahinas2012} and \cite{collins2012}.  
Through sophisticated data capture methods, such as
next generation sequencing,
researchers are able to rapidly map the complex microbiome \cite{schloss2009}.  
Nevertheless this line of research is challenging due to the shear high-dimensional nature of the microbiome, as well as the
ultra-high $p$ and small $n$ problem.  An added complexity is the manner in which the covariates are associated with
each other.
In terms of the `tree-of-life' schematic, bacterial groups at different
taxon levels have an associated phylogeny. Recent works incorporating taxonomic
information use various methods to identify important
taxonomic features, 
\cite{garcia2014},
\cite{shakar2015},
\cite{zhang2015},
\cite{liquet2016},
and
\cite{shi2016}.  Common to all these papers is the incorporation of the underlying microbiology to establish sophisticated computational schemes over and above the already demanding computational methods required to obtain the microbiome data. Due to the complexity of these methods, validation is established mainly through simulations.  
In this paper we profile a variable selection method that also incorporates the tree-of-life schema, similar in scope to the above papers,
but different in terms of how the regularization is carried out.  Extensive simulations are performed but we also obtain theoretical oracle properties
for our method, which to our knowledge, is very novel for microbiome research.

We now provide a summary of the paper.  As motivation, Section \ref{sec:background} provides the microbiology background to this paper.
Although we are addressing metagenomics through next generation sequencing of the 16S rRNA gene, the methodology is sufficiently rich to incorporate other `-omics' structures.  In Section \ref{sec:glm}
we provide the development of our penalization of the phylogenetic tree.  
An application to an infectious disease is discussed in Section \ref{sec:app}.
We include a detailed discussion because this infectious disease is currently a very important public health concern, and because it is the
main motivation behind the development of the methods in this paper.  
Further detailed simulation results are presented in Section \ref{sec:sim} with comparisons made to other
variable selection procedures. 
This is followed by Section \ref{sec:theoretical} which establishes the main theoretical properties with all of
the proofs provided in Appendix \ref{app:proofs}.

\section{A microbiology primer}
\label{sec:background}
Metagenomics, like the other `-omics', bears a structural relationship between covariates. Bacteria exhibit a tree-like relationship with each other, often violated {\it via} lateral gene transfer. As a result, their systematic taxonomy is constantly in flux; see for instance the paraphyletic {\it Clostridium} \cite{wiegel2006,rainey2009}. Thus their OTU (operational taxonomic unit) proxies, typically the 16S rRNA gene, also exhibit a tree-like structure with patterns of cycles at deep taxon classification. Our resolution depends on the length of the 16S rRNA region, the degree of lateral gene transfer, and the reliability of the reads.

In many problems, it is important to develop a means of selecting the OTUs having dominating roles in the microbial systems at hand. The relationship between OTUs is such that some of them may represent the same type of bacteria. Alternatively, some spurious OTUs are artifacts of the laboratory sequencing protocol. Rather than select OTUs on their individual merits, we want to select them based on their group affiliations. 
 
Precisely identifying and characterizing bacteria currently requires genetic and metabolic analysis of clonal colonies {\it in vitro}. In studying the composition of microbiomes {\it in vivo}, we require a general descriptor. Absent of molecular characterization, there is no general descriptor of bacterial groups, a property which constrained bacterial research for decades. 
A broader analysis of mixed bacterial communities is available, sacrificing the resolution of clonal studies. 
This involves the collection and sequencing of genes common to all bacteria, or to the group of interest.
This sequencing may be scaled up to generate
billions of sequences in tandem.  We refer to this as MPS (massively parallel sequencing). For an in depth description from a mathematical perspective, see \cite{rush2012}.

In the present paper, we consider MPS data targeting the V3-V5 region of the 16S rRNA gene, whose merits we discuss in \cite{kim2014}. Crucially, this is a gene common to all bacteria with sufficient evolutionarily conserved variability to provide a basis for defining bacterial groups \cite{woese2013}, albeit not the sole defining descriptor. MPS technologies are currently incapable of providing the full 16S sequence, so a hypervariable region is targeted, of which there are nine. Hypervariable is understood as evolutionarily conserved but more variable than the average over the entire sequence. The various hypervariable regions provide differing resolutions in different bacterial groups. An analysis of the precision of each region for identifying pathogenic bacteria is presented in \cite{chakravorty2007}. The V3-V5 regions provide estimates similar to the full 16S gene in terms of species richness \cite{youssef2009,zhao2013}, and it has been shown to provide accurate community structure and low bias estimates of some taxa \cite{bergmann2011,vilo2012}. A limitation is its poor species level resolution, as found in \cite{chakravorty2007}.

To facilitate comparison between microbiomes, sequences are grouped into OTUs according to some similarity criterion. This is a data-driven proxy to species delineation; we use OTUs in an operational manner. OTUs consist of sequences which are phylogenetically close. Phylogenetic proximity between OTUs is determined by the inter-OTU phylogenetic divergence of the constituent sequences. This provides a structural relationship between the OTUs. One manner of presenting this structure is to assign phylotypes to the OTUs, sequence-based consensus taxonomical classification of the clusters. This induces a rooted tree hierarchy between the OTUs. This structure is exploited in the present article.
 
\section{Generalized linear models and the $\Phi-$LASSO}
\label{sec:glm}
In this section, we provide details of what we call the 
phylogenetic LASSO (least absolute shrinkage and selection operator). As motivation, let us review the hierarchial H-LASSO, presented in \cite{zhou2010} in the context of penalized least-squares. Consider the linear model 
\begin{align}
	Y=X\beta+\epsilon \label{eqn:model},
\end{align}
for response $Y$, covariates $X$, parameter vector $\beta\in\mathbb R^p$, and error $\epsilon$. Suppose the covariates $X$ may be assigned to $k$ mutually exclusive groups $K_j$, $j=1,...,k$. Let $\beta_j\in\mathbb R^{p_j}$ be the subvector of $\beta$ whose coefficients correspond to the covariates in $K_j$, $|K_j|=p_j$, where $|\cdot|$ denotes set cardinality. We have $p=\sum_{j=1}^kp_j$.

We can decompose $\beta$ by $\beta_j=d_j\alpha_j$ where $d_j\geq0$ and $\alpha_j\in\mathbb R^{p_j}$. Clearly this decomposition is not unique, but this does not ultimately matter, see Lemma \ref{lem:ME} below. Let $d=(d_1,...,d_k)\in\mathbb R^k$, $\alpha=(\alpha_1,...,\alpha_k)\in\mathbb R^p$, and define $\varphi$ as the mapping given by $(d, \alpha) \mapsto \beta$.

We define the H-LASSO estimator $\hat\beta$ of (\ref{eqn:model}) {\it via} $\hat\beta=\varphi(\hat d,\hat\alpha)$ where $(\hat d,\hat\alpha)$ maximizes the penalized least squares function
$$
-\frac12\sum_{i=1}^n(Y_i-X_i\cdot\varphi(d,\alpha))^2-\lambda_1\sum_{j=1}^kd_j-\lambda_2\sum_{j=1}^k||\alpha_j||_1,
$$
where the tuning parameters $\lambda_1,\lambda_2>0$ are fixed, $\cdot$ denotes dot product, and $||\cdot||_q$ is the usual $l_q$-norm for $q > 0$. Here we penalize the groups $K_j$ by the middle term $\lambda_1\sum_{j=1}^kd_j$ and the individual coefficients by the third term $\lambda_2\sum_{j=1}^k||\alpha_j||_1$.

It is shown in \cite{zhou2010} that the penalty values $\lambda_1,\lambda_2$ redistribute geometrically so that we may replace them by a common coefficient $\lambda=\sqrt{\lambda_1\lambda_2}$. This leads to the result that their regularization is equivalent to maximizing
$$
-\frac12\sum_{i=1}^n(Y_i-X_i\cdot\beta)^2-2\lambda\sum_{j=1}^k\sqrt{||\beta_j||_1}.
$$
The H-LASSO is thus a nonconcave variant of the group LASSO, \cite{yuan2006}.

While \cite{zhou2010} directly treats least squares, their results generalize readily to arbitrary likelihood functions which possess an attractive oracle property, see Theorem 2, \cite{zhou2010}. Below, we provide the generalization to accommodate increased depth to the hierarchy by framing it in terms of a taxonomy.

\subsection{The $\Phi-$LASSO}
Our goal now is to create a hierarchical penalization scheme where there are multiple competing ways of grouping covariates. If these groupings have a nesting property, we can represent this as a tree, otherwise the graphical representation has cycles. In light of the above discussion, we use the convex log-likelihood function $\ell$ \textit{in lieu} of least squares in the sequel. We frame variable selection in terms of OTU selection in metagenomic analysis, but stress that the method easily accomodates other `-omic' data forms, be they from metabolomics, proteomics, and transcriptomics. Further, as life betrays the ability to transmit genetic information horizontally, we incoporate the option to use taxonomies as degenerate as the tree-of-life itself. In particular, this allows us to consider overlapping classifications of the variables, extending analysis to, say, active metabolic pathways of microbial systems or human tissues. We introduce some terminology and notation to help bridge the mathematics and systematics at hand.

\begin{definition}
	Let $X_1,...,X_d$ be the $d$ column vectors of the design matrix $X\in\mathbb R^{n\times p}$. A {\bf taxon} $\tau$ $($plural {\bf taxa}$)$ is a subset of the indices $I=\{1,...,p\}$ and $X_\tau=[X_j]_{j\in\tau}$ is the corresponding submatrix of $X$. A {\bf taxon level} is a collection of pairwise disjoint taxa $\tau_k, k=1,...,K$ whose union is $I$.
\end{definition}

Consider Table \ref{table:extax}, below, as an example. We find that taxa ${\rm Bacilli}=\{3,4,5,6,7,8\}$ and ${\it Enterococcaceae}=\{3,4,5\}$. The collection $\{{\rm Actinobacteria, Bacilli, Clostridia}\}$ is a taxon level.
In general, we subdivide the indices into taxa at $T+1$ taxon levels. We denote the $k$-th taxon of the $t$-th taxon level by $\tau^t$ or $\tau_k^t$.

\begin{table}[h!]
\begin{center}
{\tiny
\begin{tabular}{r ||l| l| l| l|c}
	Index	&Phylum	&	Class	&Order	&Family	&OTU\\ \hline\hline
	1	&Actinobacteria&Actinobacteria	&Bifidobacteriales		&{\it Bifidobacteriaceae}	&${\rm OTU}_1$\\ \hline
	2	&Actinobacteria	&Actinobacteria&Bifidobacteriales	&{\it Bifidobacteriaceae}	&${\rm OTU}_2$\\ \hline
	3	&Firmicutes	&Bacilli	&Lactobacillales&{\it Enterococcaceae}	&${\rm OTU}_3$\\ \hline
	4	&Firmicutes	&Bacilli	&Lactobacillales&{\it Enterococcaceae}	&${\rm OTU}_4$\\ \hline
	5	&Firmicutes	&Bacilli	&Lactobacillales&{\it Enterococcaceae}	&${\rm OTU}_5$\\ \hline
	6	&Firmicutes	&Bacilli	&Lactobacillales&{\it Lactobacillaceae}	&${\rm OTU}_6$\\ \hline
	7	&Firmicutes	&Bacilli	&Lactobacillales&{\it Lactobacillaceae}	&${\rm OTU}_7$\\ \hline
	8	&Firmicutes	&Bacilli	&Lactobacillales&{\it Lactobacillaceae}	&${\rm OTU}_8$\\ \hline
	9	&Firmicutes	&Clostridia &Clostridiale	&{\it Clostridiaceae 1	}&${\rm OTU}_9$\\ \hline
	10	&Firmicutes	&Clostridia	&Clostridiale&{\it Clostridiaceae 1}	&${\rm OTU}_{10}$\\ \hline
	11	&Firmicutes	&Clostridia	&Clostridiale&{\it Lachnospiraceae}	&${\rm OTU}_{11}$\\ \hline
	12	&Firmicutes	&Clostridia	&Clostridiale&{\it Lachnospiraceae}	&${\rm OTU}_{12}$\\ \hline
	13	&Firmicutes	&Clostridia	&Clostridiale&{\it Lachnospiraceae}	&${\rm OTU}_{13}$\\  \hline
\end{tabular}
}
	\caption{An example taxonomy generated for thirteen OTUs using five taxon levels.}\label{table:extax}
\end{center}
\end{table}

\begin{definition}
	Suppose we have a collection $\mathcal T$ of $(T+1)$ taxon levels, where the $(T+1)$-th taxon level consists of singletons, $\{\{j\}\}_{j=1}^d$. We call $\mathcal T$ a {\bf taxonomy}.
\end{definition}

There will be times where we wish to refer to those indices that belong to specific taxa at each level. We have:

\begin{definition}
	Let $L=(\tau^1,...,\tau^T)$ be a $T$-tuple where taxon $\tau^t$ belongs to taxon level $t$. We refer to $L$ as a {\bf lineage}, with associated indices $J=\cap_{t=1}^T\tau^t$. 
\end{definition}
We write $X_L$ {\it in lieu} of $X_J$ and define $|L|:=|J|$. When we wish to make it clear we are referring to a lineage's taxon at a particular taxon level, we write $L^t$, so that $L$ may be re-written $L=(L^1,...,L^T)$. 

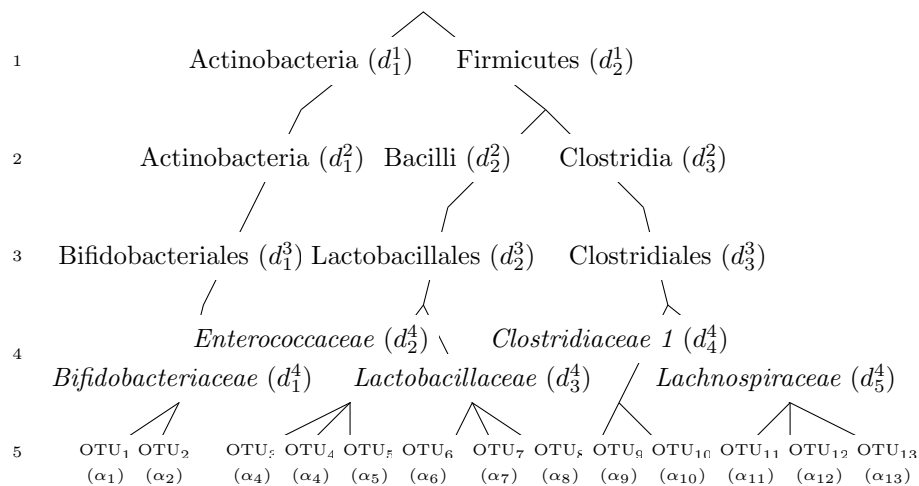
\begin{figure}[h!]
	\begin{center}
	
		\begin{tikzpicture}[scale=.65]
			\draw (-4.5,10)node[left]{\tiny 1} (-4.5,8)node[left]{\tiny 2} (-4.5,6)node[left]{\tiny 3} (-4.5,4)node[left]{\tiny 4} (-4.5,2)node[left]{\tiny 5};
			\draw (1,9)--(3.5,11)--(6,9);
			\draw (1, 9)--(0,7)--(-1,5)--(-1.5,3);
						\draw (-1.5,3)--(-3,2);
						\draw (-1.5,3)--(-2,2);
			\draw (6,9)--(4,7);
				\draw (4,7)--(3.5,5);
					\draw (3.5,5)--(2,3);
						\draw (2,3)--(0,2);
						\draw (2,3)--(1,2);
						\draw (2,3)--(2,2);
					\draw (3.5,5)--(4.5,3);
						\draw (4.5,3)--(4,2);
						\draw (4.5,3)--(5,2);
						\draw (4.5,3)--(6,2);
			\draw (6,9)--(8,7);
				\draw (8,7)--(8.5,5);	
					\draw (8.5,5)--(7.5,3);
						\draw (7.5,3)--(7,2);				
						\draw (7.5,3)--(8.5,2);				
					\draw (8.5,5)--(11,3);
						\draw (11,3)--(10,2);				
						\draw (11,3)--(11,2);				
						\draw (11,3)--(13,2);		
			\draw (1,10)node[rectangle,fill=white]{\small Actinobacteria ($d^1_1$)};		
			\draw (0,8)node[rectangle,fill=white]{\small  Actinobacteria ($d^2_1$)};		
			\draw (-1.45,6)node[rectangle,fill=white]{\small  Bifidobacteriales ($d^3_1$)};		
			\draw (-1.438,3.5)node[rectangle,fill=white]{\small  \it Bifidobacteriaceae {\rm ($d^4_1$)}};		
			\draw (-3,2)node[rectangle,fill=white]{\tiny ${\rm OTU}_1$};
			\draw (-3,1.5)node[rectangle,fill=white]{\tiny $(\alpha_1)$};		
			\draw (-1.8,2)node[rectangle,fill=white]{ \tiny ${\rm OTU}_2$};
			\draw (-1.8,1.5)node[rectangle,fill=white]{ \tiny $(\alpha_2)$};		
			\draw (6,10)node[rectangle,fill=white]{\small Firmicutes ($d^1_2$)};		
			\draw (4,8)node[rectangle,fill=white]{\small Bacilli ($d^2_2$)};		
			\draw (3.5,6)node[rectangle,fill=white]{\small Lactobacillales ($d^3_2$)};		
			\draw (1.2,4.3)node[rectangle,fill=white]{\small \it Enterococcaceae {\rm ($d^4_2$)}};		
			\draw (0,2)node[rectangle,fill=white]{ \tiny ${\rm OTU}_3$};
			\draw (0,1.5)node[rectangle,fill=white]{ \tiny $(\alpha_4)$};		
			\draw (1.2,2)node[rectangle,fill=white]{  \tiny${\rm OTU}_4$};
			\draw (1.2,1.5)node[rectangle,fill=white]{  \tiny$(\alpha_4)$};		
			\draw (2.4,2)node[rectangle,fill=white]{  \tiny${\rm OTU}_5$};	
			\draw (2.4,1.5)node[rectangle,fill=white]{  \tiny$(\alpha_5)$};		
			\draw (4.5,3.5)node[rectangle,fill=white]{\small \it Lactobacillaceae {\rm ($d^4_3$)}};		
			\draw (3.6,2)node[rectangle,fill=white]{ \tiny ${\rm OTU}_6$};	
			\draw (3.6,1.5)node[rectangle,fill=white]{ \tiny $(\alpha_6)$};		
			\draw (5.05,2)node[rectangle,fill=white]{  \tiny${\rm OTU}_7$};
			\draw (5.05,1.5)node[rectangle,fill=white]{  \tiny$(\alpha_7)$};		
			\draw (6.3,2)node[rectangle,fill=white]{  \tiny${\rm OTU}_8$};
			\draw (6.3,1.5)node[rectangle,fill=white]{  \tiny$(\alpha_8)$};		
			\draw (8,8)node[rectangle,fill=white]{\small  Clostridia ($d^2_3$)};		
			\draw (8.5,6)node[rectangle,fill=white]{\small  Clostridiales ($d^3_3$)};		
			\draw (7.3,4.3)node[rectangle,fill=white]{\small  \it Clostridiaceae 1 {\rm ($d^4_4$)}};		
			\draw (7.5,2)node[rectangle,fill=white]{  \tiny${\rm OTU}_9$};
			\draw (7.5,1.5)node[rectangle,fill=white]{  \tiny$(\alpha_9)$};		
			\draw (8.8,2)node[rectangle,fill=white]{  \tiny${\rm OTU}_{10}$};	
			\draw (8.8,1.5)node[rectangle,fill=white]{  \tiny$(\alpha_{10})$};	
			\draw (10.7,3.5)node[rectangle,fill=white]{\small \it Lachnospiraceae {\rm ($d^4_5$)}};		
			\draw (10.2,2)node[rectangle,fill=white]{ \tiny ${\rm OTU}_{11}$};
			\draw (10.2,1.5)node[rectangle,fill=white]{ \tiny $(\alpha_{11})$};		
			\draw (11.6,2)node[rectangle,fill=white]{  \tiny${\rm OTU}_{12}$};
			\draw (11.6,1.5)node[rectangle,fill=white]{  \tiny$(\alpha_{12})$};		
			\draw (13,2)node[rectangle,fill=white]{ \tiny ${\rm OTU}_{13}$};	
			\draw (13,1.5)node[rectangle,fill=white]{ \tiny $(\alpha_{13})$};		
		\end{tikzpicture}
		
	\end{center}
	\caption{A graphical representation of the taxonomy as well as
	an illustration of the decomposition of the parameters $\beta$ following the taxonomy in Table \ref{table:extax}.}\label{fig:tree}
\end{figure}

In our example, the lineages may be read directly off Table \ref{table:extax}, of which there are five. Figure \ref{fig:tree} presents the taxonomy more intuitively. Each branch of the tree represents a taxon while each horizontal row represents a taxon level, numbered 1 through 5. Each lineage is a directed path from the root to a branch.

Consider a generalized linear model with a known link function $g$, so that
$\mathbb E(Y)=g^{-1}(X\beta)$.
We decompose $\beta$ by $\beta_L=d_L\alpha_L$, where $\alpha_L\in\mathbb R^{|L|}$ as before, but now $d_L=\prod_{t=1}^Td_{L^t}$, $d_L^t\geq0$ for $t=1,...,T$. We write this decomposition as $(D,\alpha)$. Let $\varphi$ be the map $(D,\alpha)\mapsto\beta$. This decomposition is illustrated in parentheses in Figure \ref{fig:tree} below: the coefficients $d_L$ are recoverd by multiplying the terms in a lineage. We extend the optimization criterion to: 
\begin{align}
\ell(\varphi(D,\alpha);Y,X) - \sum_{t=1}^T\lambda_t\sum_{k=1}^{K^t}d_k^t -\lambda_{T+1}||\alpha||_1 \label{eqn:defn}
\end{align}
where as usual $\ell$ is the log-likelihood, and $\lambda_t>0$ for $t=1,...,T+1$. The centre double sum is the groups penalty and addresses the increased depth of our taxonomy in contrast to \cite{zhou2010} and $K^t$ is the number of taxa in taxon level $t$. When the lineage consists of a single taxon level $T=1$, this reduces to the criterion in \cite{zhou2010}.

We appear to suffer an affluence of tuning parameters. Lemma \ref{lem1}, below, shows that criterion (\ref{eqn:defn}) is equivalent to the single tuning parameter criterion 
\begin{align}\label{Q}
	\ell(\varphi(D,\alpha);Y,X) - \sum_{t=1}^T\sum_{k=1}^{K^t}d_k^t - \lambda ||\alpha||_1\ \ .
\end{align}
Let $d^t$ denote the vector of groups coefficients of the $t$-th taxon level.
Intuitively, as the tuning parameters $\lambda_t$ are redistributed, $D$ and $\alpha$ change in geometric response due to the relationship $\varphi(D,\alpha)=\beta$. Choosing $\lambda = \lambda_t$, $t=1,...,T+1$, we need only concern ourselves with one tuning parameter.

\subsection{$\Phi-$LASSO algorithm}

The algorithm used to obtain the $\Phi-$LASSO estimate relies on iterative adaptive reweighting, deriving from the reformulation of the $\Phi-$LASSO in (\ref{Q}).

Let $\psi:\beta\mapsto(D,\alpha)$ be the map from $\beta$ to the unique maximizer of (\ref{Q}) over $\varphi^{-1}(\beta)$. In this way $D$ and $\alpha$ may be viewed as projections onto the first and second elements of $\varphi^{-1}(\beta)$. Let $\Lambda$ be a positive collection of tuning parameters $\lambda$.
The pseudo-code is spelled out below.  We remark that convergence is typically achieved quickly, so that the bottlenecks are the weighted LASSO problem and calculation of $w$. 

\bigskip

\begin{algorithm}[h!]
\caption{$\Phi-$LASSO}
\label{algorithm}
\begin{algorithmic}
\Procedure{phylasso}{}
\For{each $\lambda\in\Lambda$}
\State obtain initial LASSO estimate $\hat\beta^{0(\lambda)}$
\For{$k\in\mathbb N$}
\State $w=\varphi(D(\hat\beta^{(k-1)(\lambda)}),1)$
\State solve weighted LASSO for $\hat\beta^{k(\lambda)}$ with weights $w^{-1}$
\State break if $\Delta\hat\beta^{\lambda}<$ threshold
\EndFor
\EndFor
\EndProcedure
\end{algorithmic}
\end{algorithm}

\section{Application to $\mathbf{\it  Clostridium \ difficile}$ Infection }
\label{sec:app}

{\it Clostridium difficile} ({\it C.\ difficile}) infection (CDI) is the most frequent cause of healthcare-associated 
infections and its rates are growing in the community \cite{Loo, brandt}.  
One of the major risk factors for developing CDI is use of antibiotics. The 
healthy and diverse bacteria which reside within the colon are the major defense against the growth of {\it C.\ difficile}.  
Antibiotics kill these bacteria and allow {\it C.\ difficile} to multiply, produce 
toxins and cause disease. The available treatments for this infection are the antibiotics: 
metronidazole, vancomycin and fidaxomicin.  The efficacy of these antibiotics is limited 
with high recurrence rates, \cite{Gough}.

An alternative to antibiotic 
therapy for CDI, in particular for recurrent and refractory diseases, is to infuse healthy gut bacteria directly into the 
colon of infected patients to combat {\it C.\ difficile} by a procedure known as faecal microbiota transplantation (FMT).  
FMT is a process in which a healthy 
donor's 
stool 
is administered to an affected patient.  This can be performed using
a colonoscope, nasogastric tube, or enema.  FMT serves to reconstitute the altered colonic flora, in contrast to treatment with antibiotic(s), 
which can further disrupt the establishment of key microbes essential in preventing recurrent CDI. The 
literature reveals a cumulative clinical success rate of over 90\% in confirmed recurrent CDI cases, \cite{Gough}.

There is a growing interest in the microbiome of CDI patients 
following an FMT \cite{shahinas2012, hamilton2013, song13, weingarden2014, schubert2014}.  It is noted that there are 
vast differences in:
the route of administration with all forms covered; donor selection criteria, some used family members, while others used universal donors; 
sample sizes, although all studies had small sample sizes; and sequencing procedures and equipment.  Despite these differences, there are two fundamental 
points of agreement across all studies.  The first is that CDI patients have low diversity in their microbiome, pre-FMT, and that after receiving an FMT(s), their diversity increased. The second is CDI patients who were successfully cured with FMT undergo changes in their microbiome which initially have similarities to that of their donors, see \cite{shahinas2012}.  

We will examine the microbiome data coming from a subset of the CDI-patients treated with FMT by
the second author \cite{lee2014}, covering the period 2008--2012. 

From the 17 selected patients, we note that 13 of these patients
responded to a single FMT.  We used a taxonomy describing phylum, class, order, {\it family}, and {\it genus}. 
The predictors consisted of the relative abundances of 220 pre-FMT OTUs and 347 post-FMT OTUs.
Here we consider logistic regression to model the response for two scenarios. First we wish to know whether the pre-FMT microbiome could predict a clinical response to an FMT. Second, we wish to know whether the composition of the post-FMT microbiome could anticipate the need for additional FMTs. To select the tuning parameter for the $\Phi$-LASSO, we perform leave-one-out cross-validation (LOO-CV). The optimal tuning parameter was selected by AUC (area under the curve) and BS (Brier score).
This is the
same dataset where phylum interaction was investigated, \cite{martinez2016}.

\begin{figure}
\begin{center}
\includegraphics[scale = 0.5]{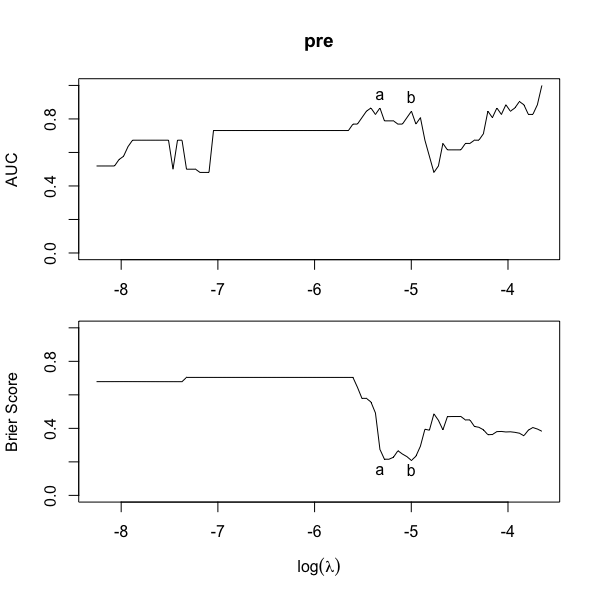}
\end{center}\caption{AUC and BS obtained by leave-one-out cross-validation for pre-FMT OTUs. Labeled are the tuning parameters by (a) AUC  ($\lambda = 0.0049$) and (b) BS
($\lambda = 0.0067$).}\label{fig:loofitpre}
\end{figure}

We are challenged by a sparse predictor matrix as well as a small sample size.  Since AUC tends to assign a perfect score to the null model by the way it handles ties, we incorporate BS to compensate for this
affect.
The results of logistic regression using the pre-FMT OTUs as covariates are captured in Figure \ref{fig:loofitpre}. Using the globally optimal BS of 0.208, the corresponding tuning parameter is
$\lambda = 0.0067$.  Relative to the BS,
the locally optimal AUC value is 0.846, with the corresponding tuning parameter of $\lambda = 0.0049$.
The globally optimal AUC value is 0.865 which would be associated with the null model.  
Below the figure in Table \ref{tab:loofitpre}, we display the family and genus of the selected OTUs.  Using LOO-CV, the frequency along with the averaged estimate and LOO-CV 
standard errors are reported.  Due to the small sample size the variability in the parameter estimates are large which is to be expected.  Concentrating on the frequency using LOO-CV
we notice that OTU 7 associated with the family {\it Lactobacillaceae} and the
corresponding genera {\it Lactobacillus} seem to have some positive predictability.  This appears to be consistent with some earlier findings that the last author was involved in using a different FMT
patient cohort, \cite{shahinas2012}.  We note that some family/genera identification appear to be the same for different OTUs.  This in interpreted to mean differences at the species taxon.

\begin{figure}
\begin{center}
\includegraphics[scale = 0.5]{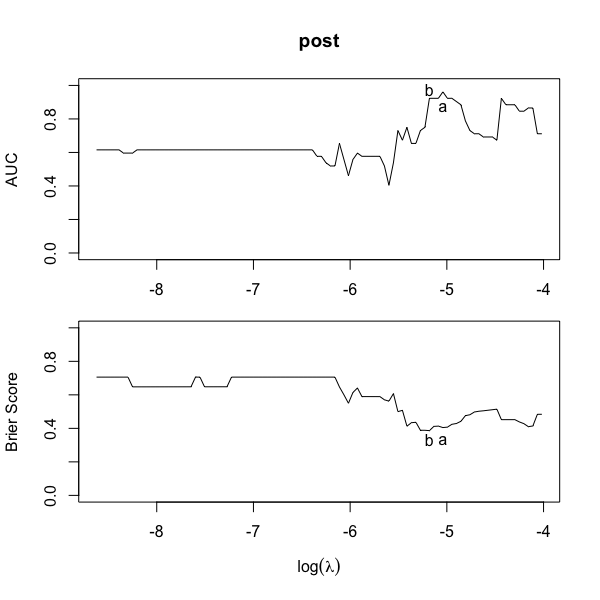}
\end{center}\caption{AUC and BS obtained by leave-one-out cross-validation for post-FMT OTUs. Labeled are the tuning parameters by (a) AUC  ($\lambda = 0.0065$) and (b) BS
($\lambda = 0.0056$).}\label{fig:loofitpost}
\end{figure}

For the post-FMT OTUs, we obtain a globally optimal BS of 0.386,  $\lambda = 0.0056$, with corresponding AUC value 0.923, $\lambda = 0.0065$. The globally optimal AUC is 0.961.  This
is displayed in Figure \ref{fig:loofitpost}.  In Table \ref{tab:loofitpost}, we display the family and genus of the selected OTUs.  Using LOO-CV, the frequency along with the averaged estimate and LOO-CV 
standard errors are reported.  Again due to the small sample size the variability in the parameter estimates are large.  Concentrating on the frequency using LOO-CV
we notice that OTU 5 associated with the family {\it Enterococcaceae} and the
corresponding genera {\it Enterococcus}, as well as OTU 17  corresponding to the family {\it Bacteroidaceae} and genera {\it Bacteroides} appear to have some positive predictability.  
This again is consistent with some earlier results, \cite{shahinas2012}.  

We realize that the interpretation of these results from a bacteriological perspective is well beyond the scope of this medium, and the purpose
of this section is to present a portrayal of the type of statistical analysis that comes from looking at the microbiome as a covariate.  We 
highlight above some of the more obvious interpretations that is consistent with what has been observed in other studies.  Indeed,
results of this work will interest researchers and companies working toward refinement of FMT, {\it via} establishment of central stool banks or creation of synthetic stool. With results
obtained using larger sample sizes it will be possible to select 
species that come from the genera identification and culture them in a laboratory biochemistry setting.  Thus one could
strategically pool samples to achieve a desired composition, to supplement stool with specific microorganisms or to be able to prepare recipient-specific synthetic stool as
{\it per} the RePOOPulate project, \cite{petrof2013}.
At this point however, we will end this discussion with the comment that a more in depth
metageonomic analysis of a recent clinical trial, \cite{lee2016}, is under investigation. 

\begin{table}
\begin{center}
{\tiny
\begin{tabular}{|c|ll|cr|cr|} \hline
& \multicolumn{2}{|c|}{Phylogeny} &  \multicolumn{2}{|c|}{$\lambda=0.0049$}& \multicolumn{2}{|c|}{$\lambda=0.0067$}\\ 
OTU & Family & Genus & Frequency & $\hat\beta({\rm SE})$ & Frequency & $\hat\beta({\rm SE})$ \\\hline\hline
1 & {\it Enterobacteriaceae} & {\it Klebsiella} & 0.24 & 10.8 (41.6) & 0.35 & -4.65 (13.7) \\ \hline
2 & {\it Enterobacteriaceae} & {\it Escherichia/Shigella} & 0.59 & -49.4 (122) & 0.88 & -107 (143) \\ \hline
3 & {\it Streptococcaceae} & {\it Streptococcus} & 0.82 & -304 (323) & 0.76 & -395 (425) \\ \hline
4 & {\it Lachnospiraceae} & {\it Blautia} & 0.82 & 239 (281) & 0.18 & 20.4 (66.6) \\ \hline
5 & {\it Enterococcaceae} & {\it Enterococcus} & 0.47 & -94.9 (137) & 0.65 & -164 (264) \\ \hline
6 & {\it Lactobacillaceae} & {\it Lactobacillus} & 0.53 & -15.5 (25.2) & 0.82 & -29.8 (29.6) \\ \hline
7 & {\it Lactobacillaceae} & {\it Lactobacillus} & 0.88 & 349 (335) & 0.88 & 387 (415) \\ \hline
11 & {\it Lachnospiraceae} & unclassified & 0.53 & -60.9 (77.5) & 0.59 & -68.6 (85.1) \\ \hline
13 & {\it Veillonellaceae} & {\it Veillonella} & 0.29 & 78.5 (127) & - &  - \\ \hline
24 & {\it Veillonellaceae} & unclassified & 0.65 & -108 (141) & - & - \\ \hline
26 & {\it Veillonellaceae} & unclassified & - & - & 0.47 & -97.9 (142) \\ \hline
30 & {\it Veillonellaceae} & {\it Veillonella} & 0.76 & 176 (258) & - & - \\ \hline
33 & {\it Veillonellaceae} & {\it Veillonella} & - & - & 0.53 & 55.6 (107) \\ \hline
\end{tabular}
}
\caption{The selected OTUs for pre-FMT microbiomes at the tuning parameters selected by AUC ($\lambda = 0.0049$) and BS ($\lambda = 0.0067$) leave-one-out cross-validation on relative abundances.}\label{tab:loofitpre}
\end{center}
\end{table}

\newpage

\begin{table}[h!]
\begin{center}
{\tiny
\begin{tabular}{|c|ll|cr|cr|} \hline
& \multicolumn{2}{|c|}{Phylogeny} &  \multicolumn{2}{|c|}{$\lambda=0.0065$}& \multicolumn{2}{|c|}{$\lambda=0.0056$}\\ 
OTU & Family & Genus & Frequency & $\hat\beta({\rm SE})$ & Frequency & $\hat\beta({\rm SE})$ \\\hline\hline
1 & {\it Enterobacteriaceae} & {\it Klebsiella} & 0.94 & 7.79 (28.8) & 0.82 & 8.65 (28.9) \\ \hline
2 & {\it Enterobacteriaceae} & {\it Escherichia/Shigella} & 0.82 & -17.6 (80.7) & 0.94 & -17.1 (80.9) \\ \hline
3 & {\it Streptococcaceae} & {\it Streptococcus} & 0.88 & -22.6 (42.2) & 0.94 & -52.4 (82.3) \\ \hline
5 & {\it Enterococcaceae} & {\it Enterococcus} & 0.94 & 98.5 (261) & 0.94 & 147 (275) \\ \hline
17 & {\it Bacteroidaceae} & {\it Bacteroides} & 0.94 & 109 (334) & 0.94 & 111 (333) \\ \hline
21 & {\it Acidaminococcaceae} & {\it Acidaminococcus} & 0.65 & -1.34 (1.85) & 0.71 & -15.6 (33.9) \\ \hline
\end{tabular}
}
\caption{The selected OTUs for post-FMT microbiomes at the tuning parameters selected by AUC ($\lambda = 0.0065$) and BS ($\lambda = 0.0056$) leave-one-out cross-validation on relative abundances.}\label{tab:loofitpost}
\end{center}
\end{table}

\break

\section{Simulations}
\label{sec:sim}
\setcounter{equation}{0}
In this section, we report on simulations that approximate the covariance structure we may see in practice with respect to phylogenetic proximity of the OTUs.  We will
consider $T=5$ taxon levels where each level is balanced and grows according to $4^k$, $k=0,1,2,3,4,5,6$.  In taxonomy language we will consider a
balanced taxonomy where we have a 
single `phylum' ($k=0$) followed by `class' ($k=1$), `order' ($k=2$), `{\it family}' ($k=3$),  `{\it genus}' ($k=4$), and `{\it species}' ($k=5$).  This will 
then lead to 4096 OTUs ($k=6$).

\begin{figure}[h!]
\begin{center}
\includegraphics[scale=.5]{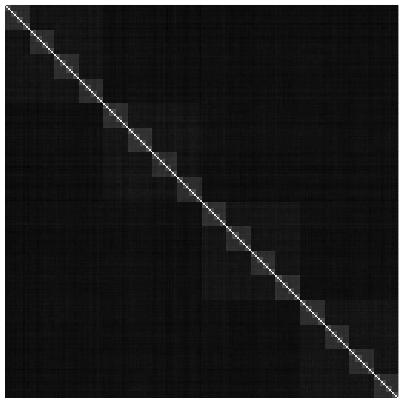}
\caption{Heatmap of covariance matrix for 10,000 point validation set from the tuning parameter step. Displayed is the submatrix corresponding to a single `class'.}\label{fig:covheat}
\end{center}
\end{figure}

To each taxon we introduce taxon-wise covariation. This covariance structure is presented in Figure \ref{fig:covheat} for a single `class' (1024 OTUs) using a 10,000 point sample. The true parameters in the simulation are represented by two classes, with one class dominating the other 3:1.

\begin{figure}
\begin{center}
\begin{tikzpicture}[scale=.5]
	\draw (0,5.5)--(-1,5);
		\draw (-1,5)--(-1.5,4);
			\draw (-1.5,4)--(-2.5,3);
				\draw (-2.5,3)--(-3.5,2);
					\draw (-3.5,2)--(-4.5,1);
						\draw (-4.5,1)--(-4.8,0);
						\draw (-4.5,1)--(-4.6,0);
						\draw (-4.5,1)--(-4.4,0);
						\draw (-4.5,1)--(-4.2,0);
					\draw (-3.5,2)--(-3.5,1);
						\draw (-3.5,1)--(-3.8,0);
						\draw (-3.5,1)--(-3.6,0);
						\draw (-3.5,1)--(-3.4,0);
						\draw (-3.5,1)--(-3.2,0);					
					\draw (-3.5,2)--(-2.5,1);
						\draw (-2.5,1)--(-2.8,0);
						\draw (-2.5,1)--(-2.6,0);
						\draw (-2.5,1)--(-2.4,0);
						\draw (-2.5,1)--(-2.2,0);			
				\draw (-2.5,3)--(-1.5,2);
					\draw (-1.5,2)--(-1.5,1);
						\draw (-1.5,1)--(-1.8,0);
						\draw (-1.5,1)--(-1.6,0);
						\draw (-1.5,1)--(-1.4,0);
						\draw (-1.5,1)--(-1.2,0);		
			\draw (-1.5,4)--(-.5,3);
				\draw (-.5,3)--(0,2);
					\draw (0,2)--(-.5,1);
						\draw (-0.5,1)--(-0.8,0);
						\draw (-0.5,1)--(-0.6,0);
						\draw (-0.5,1)--(-0.4,0);
						\draw (-0.5,1)--(-0.2,0);					
					\draw (0,2)--(.5,1);
						\draw (0.5,1)--(0.8,0);
						\draw (0.5,1)--(0.6,0);
						\draw (0.5,1)--(0.4,0);
						\draw (0.5,1)--(0.2,0);
	\draw (0,5.5)--(1,5);
		\draw (1,5)--(1.5,4);
			\draw (1.5,4)--(1.8,3);
				\draw (1.8,3)--(1.8,2);
					\draw (1.8,2)--(1.3,1);
						\draw (1.3,1)--(1.0,0);
						\draw (1.3,1)--(1.2,0);
						\draw (1.3,1)--(1.4,0);
						\draw (1.3,1)--(1.6,0);
					\draw (1.8,2)--(2.3,1);
						\draw (2.3,1)--(2.0,0);
						\draw (2.3,1)--(2.2,0);
						\draw (2.3,1)--(2.4,0);
						\draw (2.3,1)--(2.6,0);
\end{tikzpicture}
\end{center}
\caption{Pruned taxonomy, from an initial 1024 leaves.}\label{fig:sim5}
\end{figure}
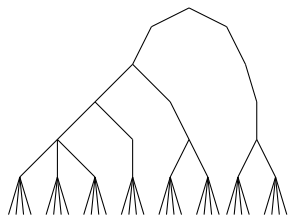

We consider random Gaussian data generated from a 4096 covariate sparse linear model where 32 covariates have corresponding parameter $\beta_j=2$ and the rest being zero. We present the corresponding pruned tree in Figure \ref{fig:sim5}. 

The species information is deliberately lost to simulate uncertainty in species assignment to OTUs, which mimics the current technology limitation in 16s rRNA sequencing, see \cite{chakravorty2007}.  Thus the deepest taxon level used in the fit is the genus level. We introduce a class, order, family, genus, and species-wise covariation $W_{c}$, $W_{o}$, $W_{f}$,  $W_{g}$, and $W_{s}$, samples from normal distributions $N(0,0.5^2)$, $N(0,1^2)$, $N(0,2^2)$, $N(0,3^2)$, $N(0,4^2)$, respectively, where
$N(0,\sigma^2)$ denotes a normal distribution with mean 0 and variance $\sigma^2$. The predictors used are then defined as $X=(Z+W_p+W_c+W_f+W_g+W_s)/\sqrt{55.25}$ where $Z\sim N_{4096}(0,5^2 I_{4096})$ is a 4096 multivariate normal distribution with mean vector 0 and covariance matrix $I_{4096}$, the $4096 \times 4096$ identity matrix. 

\subsection{Tuning parameter selection}\label{ssec:tun}

To select the tuning parameter for each $(n,\sigma)$ pair, we replicate 100 data sets $Y_i\sim N(X_i\beta,\sigma)$, $1\leq i\leq n$. We select the tuning parameter $\lambda^{(n,\sigma)}$ minimizing across all models the MSPE (mean squared prediction error) for an independent 10,000 data point validation set generated from the same distribution. We also fit SCAD models using the same datasets, using the same validation set to select the appropriate tuning parameter.

\subsection{Performance}

To evaluate performance of our models, we consider four measures for 100 replicates: SSE (sum of squared error), MSPE, `recall', and `precision'. Recall is defined as the proportion of covariates correctly selected relative to true parameters, {\tt tp/(tp+fn)}, and precision is defined as the proportion of covariates correctly selected relative to total covariates selected, {\tt tp/(tp+fp)}, where {\tt tp} is true positive, {\tt fn} is false negative and {\tt fp} is false positive.

We compare the performance of the $\Phi$-LASSO to SCAD. We also consider OLS (ordinary least squares) for the oracle model where the exact $\beta_j=0$ is known. MSPE for performance is evaluated using a new 10,000 data point validation set generated independently of the tuning validation set.

\subsection{Results: Tuning}

\begin{figure}
\begin{center}
\includegraphics[scale=.8]{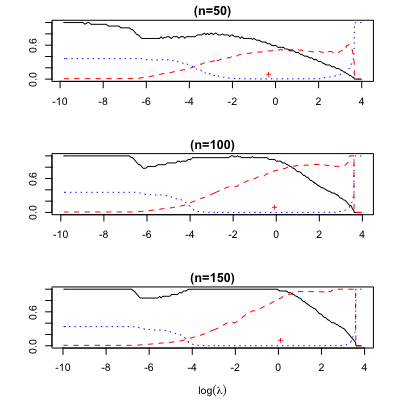}
\caption{Median recall (solid) and precision (dashed) for $\Phi$-LASSO against $\log(\lambda)$ for sample sizes $n=50,100,150$. Included is the median MSPE curve scaled by largest median (dotted). The `$+$' indicates the chosen ($\log$) tuning parameter.}\label{fig:tune5_3}
\end{center}
\end{figure}

Figure \ref{fig:tune5_3} presents recall and precision against the tuning parameters. Also displayed is the MSPE curve, scaled to the unit interval $[0,1]$. Note that the selected parameters $\lambda$ lie in the middle of a relatively flat MSPE region. Perturbations of $\lambda$ would not overly affect predictive performance. The same holds with respect to SSE (not shown). The shapes of the MSPE curves agree in large measure, aside from the null models, which have lower SSE than the sparsest estimators. Recall and precision are at odds, where for low sample size one must be sacrificed against the other.  
The selection of $\lambda$ favours higher recall over precision; the coefficients for false positives may be very small and hence affect prediction in a minor way, but false negatives are a complete loss of a structural signal. There is an interesting dip in recall for sample sizes $n\geq100$ in the region preceding the low, stable MSPE region where the estimators overfit the data. The clear separation of these two regions reflects well on the $\Phi-$LASSO stability.

\subsection{Results: Performance}

The results for SSE and MSPE are presented in Table \ref{tab:error}. SCAD struggles with the covariance structure, performing poorly in estimation. Incredibly, it appears to perform well in MSPE. While SCAD struggles to recall the correct OTUs due to high correlation within species, to its credit it is able to select some related taxa, where the covariance structure leads to similar predictive performance but more flexibility in fitting the model. The $\Phi$-LASSO exhibits none of the SCAD's difficulties. It quickly converges to agreement with the oracle estimator in estimation and prediction.

The $\Phi$-LASSO quickly approaches near perfect recall as sample size increases whereas SCAD becomes stalled at 25\%. The $\Phi$-LASSO consistently improves in precision, dropping false positives. For SCAD, after an initial improvement in precision, it drops as it selects incorrect but related OTUs. When we present the $\Phi$-LASSO and SCAD estimates with a validation set consisting of uncorrelated covariates, there is negligible change in $\Phi$-LASSO performance, while the SCAD's predictive performance matched its poor estimation performance.

\subsection{Additional simulations}

We comment on some additional simulations. In the low dimensional, $p < n$, regime with simple sample-wise covariation structure, SCAD performs moderately better than the $\Phi$-LASSO, with the adaptive LASSO trailing behind. This is consistent across sample size and noise level. For alternative $p >n$ scenarios ($p=1000$), the $\Phi$-LASSO performs well, although experiences some difficulty with precision for $n=250$. This is qualitatively different than the issue with SCAD in the previous section, as recall is nearly perfect. The false positives generally correspond to relatively small coefficients,  so that it appears to result from early exit from the fitting algorithm.

\begin{table}[h!]
\begin{center}
\begin{tabular}{ r|r|r|r|r|r|r }\hline
& \multicolumn{3}{|c|}{SSE} &  \multicolumn{3}{c}{MSPE} \\
$n$ & \multicolumn{1}{c}{OLS} & \multicolumn{1}{c}{$\Phi$-LASSO} & \multicolumn{1}{c|}{SCAD} & \multicolumn{1}{c}{OLS} & \multicolumn{1}{c}{$\Phi$-LASSO} & \multicolumn{1}{c}{SCAD}  \\ \hline
50& 88.10(46.2) & 167.00(62.50) & 621.00(230.00) & 3.20(1.21) & 14.80(17.2) & 167.00(162.00) \\
100& 19.80(6.05) & 37.70(12.00) & 384.00(5.28) & 1.47(0.14) & 2.43(0.48) & 9.24(0.56) \\
150& 11.10(3.69) & 15.90(6.14) & 383.00(4.47) & 1.25(0.07) & 1.49(0.18) & 8.66(0.38) \\
200& 8.05(2.48) & 10.70(4.29) & 379.00(4.35) & 1.18(0.05) & 1.29(0.10) & 8.40(0.24) \\
250& 6.25(1.77) & 7.54(2.45) & 380.00(3.36) & 1.13(0.03) & 1.19(0.06) & 8.27(0.20) \\ \hline
\end{tabular}
\caption{Estimation and prediction error for the oracle estimator (OLS), $\Phi$-LASSO, and SCAD. Presented are the mean error (standard error).}\label{tab:error}
\end{center}
\end{table}

\begin{table}
\begin{center}
\begin{tabular}{ r|r|r|r|r }\hline
& \multicolumn{2}{|c|}{Recall} &  \multicolumn{2}{c}{Precision} \\
$n$ & \multicolumn{1}{c}{$\Phi$-LASSO} & \multicolumn{1}{c|}{SCAD} & \multicolumn{1}{c}{$\Phi$-LASSO} & \multicolumn{1}{c}{SCAD}  \\ \hline
50 & 0.64(0.11) & 0.18(0.07) & 0.52(0.12) & 0.50(0.34) \\
100 & 0.92(0.04) & 0.25(0.00) & 0.71(0.09) & 0.93(0.12) \\
150 & 0.98(0.02) & 0.25(0.00) & 0.86(0.07) & 0.75(0.16) \\
200 & 0.99(0.01) & 0.25(0.00) & 0.92(0.06) & 0.54(0.11) \\
250 & 0.99(0.01) & 0.25(0.00) & 0.95(0.04) & 0.56(0.11) \\ \hline
\end{tabular}
\caption{Recall and precision for the $\Phi$-LASSO and SCAD. Presented are the mean  (standard error).}\label{tab:error}
\end{center}
\end{table}

\pagebreak

\section{Theoretical Results}
\setcounter{equation}{0}
\label{sec:theoretical}

In this section, we consider the general objective function, 
$$Q^*(\lambda_1,\dots, \lambda_{T+1}, D, \alpha) = \ell(\varphi(D,\alpha)) - \sum_{t=1}^T \lambda_t ||D_t||_q^q - \lambda_{T+1}||\alpha||_q^q$$ where $\ell$ is the known but arbitrary log-likelihood, the remainder is the penalization, and $\|\cdot\|_q$  is the $l_q$-norm, $q>0$.  We will also make use of the notation `$'$' for matrix transpose, `$\ll$',  `$\ll_p$' to mean `big oh' and `big oh in probability', respectively, $o$, $o_p$ to mean `little oh' and `little oh in probibility', respectively, 
`$\asymp$' to mean the ratio of two sequences converge to a positive constant, `$\leadsto$' to mean convergence in distribution, and `$\wedge$' to mean minimum.  All proofs are provided in Appendix \ref{app:proofs}.

The first result allows us to consider a single penalty parameter, that is, set $\lambda_t=1$ for all $t=1,...,T$. Let $Q_1=Q^*(\lambda_1,...,\lambda_{T+1},\,\cdot\,,\,\cdot\,)$, $Q_2=Q^*(1,...,1,\lambda_{T+1}\prod_{t=1}^T\lambda^{1/q},\,\cdot\,,\,\cdot\,)$.

\begin{lemma}[Equivalence of optimization]\label{lem1}
	Let $\lambda_t>0$, $t=1,...,T+1$ be fixed. Then $(D^1,\alpha^1)$ is a local maximizer of $Q_1$ if and only if $(D^2=(\lambda_t^{1/q}d_t^1), \alpha^2=\alpha^1/\prod_{t=1}^T\lambda_t^{1/q})$ is a local maximizer of $Q_2$ and hence $\varphi(D^1,\alpha^1)=\varphi(D^2,\alpha^2)$.
\end{lemma}
The proof is similar to that of Lemma 1 in \cite{zhou2010}, where least-squares with $l_1$-regularization is considered. 

In the sequel we assume that $\lambda_t=\lambda$ for all $t=1,...,T+1$. This next result provides a relationship between the individual effects parameters and the group coefficients. Intuitively, the mass distributes itself geometrically to minimize the penalty terms.

\begin{lemma}[Mass Equilibrium]\label{lem:ME}
	Let $(D,\alpha)\in\varphi^{-1}(\beta)$. Then $(D,\alpha)$ is a global maximizer of $Q^*$ over $\varphi^{-1}(\beta)$ if and only if $d_{\tau_k^t} = \sum_{L:L^t=\tau_k^t}||\alpha_L||_q^q=||\alpha_{\tau_k^t}||_q^q$. 
\end{lemma}
This conservation of mass result immediately leads to the following.
\begin{corollary}
	The relationship $d_{\tau}^q = \sum_{L:L^t=\tau}||\alpha_L||_q^q$ from {\rm Lemma \ref{lem:ME}} identifies the unique maximizer of $Q^*$ over $\varphi^{-1}(\beta)$.
\end{corollary}
Thus $\beta_j$ is an order $T+1$ polynomial in terms of $\alpha_1,...,\alpha_{p}$.

\begin{definition}
	Let $\psi$ be the map $\beta\mapsto(D,\alpha)$ where $(D,\alpha)$ is the unique optimizer of $Q^*|_{\varphi^{-1}(\beta)}$. We refer to $\psi$ as the {\bf partial inverse}.
\end{definition}

\begin{corollary}\label{cor:phi}
Suppose the taxonomy has two levels, $T=1$. Then the partial inverse $\psi(\beta)=(D,\alpha)$ is characterized by: if $\beta_L=0$ then $d_L=0$ and $\alpha_L=0$, if $\beta_L\neq0$ then $d_L=\sqrt{||\beta_L||_q}$ and $\alpha_L={\beta_L}/{\sqrt{||\beta_L||_q}}$.
\end{corollary}
Corollary \ref{cor:phi} is a generalization of Theorem 1 in \cite{zhou2010}.

\subsection{LAN conditions and the Oracle}

To ensure local asymptotic normality of the MLE for diverging number of parameters, we adopt the regularity conditions from \cite{lehman2003} as in \cite{zhou2010}:

\begin{itemize}
	\item[(A1)] For all $n$, the observations $(X_i,Y_i)$, $i=1,...n$, are independently and identically distributed according to the density $f_n(X_i,Y_i;\beta_n)$, where $f_n$ has common support and the model is identifiable. Further, 
$$\mathbb E_{\beta_n}\left[\frac{\partial\log f_n}{\partial\beta_{nL_j}}\right]=0\qquad\forall L,\;j=1,...,|L|$$
$$\mathcal I_{L_jK_k}(\beta_n)=\mathbb E_{\beta_n}\left[\frac{\partial}{\partial\beta_{nL_j}}\log f_n\cdot\frac{\partial}{\partial\beta_{nK_k}}\log f_n\right]=-\mathbb E\left[\frac{\partial^2}{\partial\beta_{nL_j}\partial\beta_{nK_k}}\log f_n\right].$$
	\item [(A2)]The Fisher information matrix $\mathcal I(\beta_n)=(\mathcal I_{L_jK_k}(\beta_n))$ is positive definite with bounds
$$0<C_1<\min\sigma(\mathcal I(\beta))\leq\max\sigma(\mathcal I(\beta))<C_2<\infty$$ where $\sigma(X)$ is the point spectrum of a matrix $X$,
$$\mathbb E_{\beta_n}\left[\frac{\partial}{\partial\beta_{nL_j}}\log f_n\cdot \frac{\partial}{\partial\beta_{nK_k}}\log f_n\right]^2<C_3<\infty$$
$$\mathbb E_{\beta_n}\left[\frac{\partial^2}{\partial\beta_{nL_j}\partial\beta_{nK_k}}\log f_n\right]^2<C_4<\infty.$$
	\item[(A3)] There exists an open subset $\omega_n\subset\Omega_n\subset\mathbb R^{P_n}$, $P_n$ the number of parameters, containing the true parameter point $\beta_n$ such that for almost all $(X_i,Y_i)$, $\frac{\partial^3f_n}{\partial\beta_{L_j}\partial\beta_{K_k}\partial\beta_{J_l}}$ is defined for all $\beta\in \omega_n$. Further, there exist functions $M_{nL_jK_kJ_l}$ such that $$\left|\frac{\partial^3f_n}{\partial\beta_{L_j}\partial\beta_{K_k}\partial\beta_{J_l}}\right|\leq M_{nL_jK_kJ_l}(X_i,Y_i)$$ for all $\beta\in \omega_n$ and $$\mathbb E_{\beta_n}\left[M_{nL_jK_kJ_l}^2(X_i,Y_i)\right]<C_5<\infty \ \ .$$ 
\end{itemize}
We call the conditions (A1)-(A3) as the LAN conditions and note that they are not excessively restrictive and falls within the usual framework of this type of analysis.

Consider the parameters $\beta_n\in\mathbb{R}^{p_n}$ where $p_n$ grow with $n$, and their partial inverse $\psi(\beta_n)=(D_n,\alpha_n)$. Let $w_j(\beta_n)=1$ if $\phi_j(D_n(\beta_n),1)=0$, $\phi_j(D_n(\beta_n),1)$ otherwise. Then assuming the relationship from Lemma \ref{lem:ME} we have the following equivalent expression,
\begin{align}
Q_n(\lambda_n,\beta_n)=&Q^*(n\lambda_n/(T+1),...,n\lambda_n/(T+1),D_n,\alpha_n) \nonumber \\
= &\ell_n(\beta_n)-n\lambda_n\sum_{L}\frac{||\beta_{n,L}||}{w_{L_1}}.
\label{eqn:Qn}
\end{align}
For convenience we will write $D_n(\beta_n)$ for the projection of $\psi(\beta_n)$ onto $D_n$.

We have the following consistency results for the $\Phi-$LASSO.

\begin{theorem}[Consistency]\label{main:consistency}
	Assume that the distribution satisfies the {\rm LAN} conditions. If $p_n^{4}=o( n)$ and $\lambda_n\ll n^{-1/2}$, then there exists a $\gamma_n$-consistent local maximizer $\hat\beta_n$ of $Q_n$, where $\gamma_n=\sqrt{p_n}(n^{-1/2}+\lambda_n)$. \end{theorem}
A simple choice is $\lambda_n\asymp n^{-1/2}.$ We find this provides us with taxon selection consistency.

Define $\delta_j:\mathbb R^{p_j}\to\{0,1\}$ by $\delta_j(x)=0$ if $x=0$ and 1 otherwise. Let $\delta=(\delta_1,...,\delta_k)$.

\begin{definition}
	Let $\hat\beta_n$ be an estimator for $(\ref{eqn:Qn})$. Then $\hat\beta_n$ is said to be consistent in group selection if $\mathbb P\{\delta(\hat\beta_n)=\delta(\beta_n)\}\to1$ as $n\to\infty$. Moreover, if $p_j=1$ for all $j$, the estimator is {\bf model selection consistent}.
\end{definition}

Consider $\beta^0_n\left({\hat \beta}_n\right)$ the true parameter (estimator) in $(\ref{eqn:Qn})$.  Suppose $\mathcal{A}_n=\left\{\beta^0_{n,kj} \neq 0\right\}$ and define
$\beta^0_{n\mathcal{A}_n}\left({\hat \beta}_{n\mathcal{A}_n}\right)$ to be the restriction of $\beta^0_n\left({\hat \beta}_n\right)$ to the set $\mathcal{A}_n$.

\begin{definition}\label{defn:oracle}
Let $\hat\beta_n$ be a model selection consistent estimator of $\beta_n^0.$ Suppose $\hat\beta_n$ satisfies 
$$\sqrt nA_n{\mathcal I}_n^{1/2}(\beta_{n\mathcal A_n}^0)(\hat\beta_{n\mathcal A_n}-\beta_{n\mathcal A_n}^0)\leadsto N(0,\Sigma)$$ where $A_n$ is an $r\times|\mathcal A_n|$ matrix such that $A_nA_n'\to\Sigma$, a positive semidefinite matrix, and ${\mathcal I}_n(\beta_{n}^0)$ is the Fisher information evaluated at $\beta_n^0$, as $n \to \infty$.  Then $\hat\beta_n$ is said to have the {\bf oracle property}.
\end{definition}

We have the following.

\begin{theorem}[Taxon selection consistency]\label{main:grpsel}
	Assume that the distribution satisfies the {\rm LAN} conditions. If $p_n^{(T+2)\wedge4}=o( n)$ and $\lambda_n\asymp n^{-1/2}$, then there exists a $\sqrt{n/p_n}$-consistent local maximizer $\hat\beta_n$ of $(\ref{eqn:Qn})$ so that $\lim_{n\to\infty}\mathbb P\{\delta(\hat\beta_{nL})=\delta(\beta_{nL}^0)\}=1$, the group sparsity property.
\end{theorem}

For the case of two taxon levels, $T=1$, the above result is Theorem 2, \cite{zhou2010}.
Using an adaptive modification to their loss function analogous to \cite{zou2006}, \cite{zhou2010} obtains the full oracle property. 
We obtain the oracle property in an alternative manner, by a simple modification of the taxonomy of the variables.  The full proofs are quite involved and are
relegated to Appendix \ref{app:proofs}.
Since we have now shown that consistent taxon selection holds for arbitrary number of taxon levels, we have the full oracle property in the following result using a simple modification of our taxonomy. This is possible as we are able to incorporate $T$, arbitrary, taxon levels.

\begin{theorem}[Oracle Property]\label{main:oracle}
	In addition to the {\rm LAN} conditions, suppose that the taxonomy contains an additional taxon level identical to the singleton level and $p_n^5=o(n)$. Then $\hat\beta_n$ is a $\sqrt{n/p_n}$-consistent local maximizer of $(\ref{eqn:Qn})$ satisfying  
\begin{itemize}
\item[{\rm (i)}] the sparsity property $\mathbb P \left\{\delta (\hat\beta_{n\mathcal A_n})=\delta (\beta_{n\mathcal A_n}^0)\right\}
\to1$ as $n\to\infty$, 
\item[{\rm (ii)}] asymptotic normality $$\sqrt{n}A_n\mathcal I_n^{1/2}(\beta_{n\mathcal A_n}^0)(\hat\beta_{n\mathcal A_n}-\beta_{n\mathcal A_n}^0)\leadsto N(0,\Sigma),$$
where $A_n$ is an $r \times {\mathcal A}_n$
matrix such that $A_nA_n'\to\Sigma$, a positive semidefinite matrix, and $\mathcal I_n(\beta_n^0)$ is the Fisher information matrix evaluated at $\beta_n^0$,
as $n\to\infty$.
\end{itemize}
\end{theorem}

A simple consequence of Theorem \ref{main:oracle} is that all estimators obtained with an $l_{1/q}$ penalty, $q\in\mathbb N$, $q>1$, have the oracle property.
\begin{corollary}
	Consider the estimator $\hat\beta_n=\arg\max_\beta\{\ell_n(\beta)-\lambda_n||\beta||_{1/{T+1}}^{1/{T+1}}\}$. 	Then $\hat\beta_n$ has the oracle property.
\end{corollary}
This result follows immediately if we consider a taxonomy in which all taxa are singletons.

\begin{remark}Alternatively, we can obtain the oracle property from the following modification of the objective function,
$$
\ell(\varphi(D,\alpha);Y,X) - \sum_{t=1}^T\sum_{k=1}^{K^t}d_k^t - \lambda||\alpha||_{\frac12}^{\frac12},
$$
where we have replaced the {\rm LASSO} penalty on $\alpha$ by a Bridge penalty with $\gamma=1/2$. The $l_{1/2}$-penalty yields an estimator equivalent to that obtain in {\rm Theorem \ref{main:oracle}}.
\end{remark}

\appendix

\section{Appendix:  Proofs}
\label{app:proofs}
\setcounter{equation}{0}
 In this appendix we provide all proofs.

\noindent {\bf Proof of Lemma \ref{lem1}.}  Clearly 
\beq\label{6.1}
Q_1((d_{\tau^t}),\alpha) = Q_2((\lambda_t^{\frac1q}d_{\tau^t}),\alpha/\Pi_{t=1}^T\lambda_t^{\frac1q}).
\eeq

\noindent$(\Longrightarrow)$ Let $(D^1,\alpha^1)$ be a local maximizer of $Q_1$. Thus there exists a $\delta>0$ such that if $(D',\alpha')$ satisfies $\sum_t||d_t'-d_t^1||_p+||\alpha'-\alpha^1||_q<\delta$, then $Q^*(D',\alpha')\leq Q^*(D^1,\alpha^1)$.
By the identity (\ref{6.1}), 
$$
	Q_2(D''=(\lambda_t^{\frac1q}D_{\tau^t}'),\alpha''=\alpha'/\Pi_{t=1}^T\lambda_t^{\frac1q}) = Q_1(D',\alpha')
				\leq Q_1(D^1,\alpha^1)
				\leq Q_2(D^2,\alpha^2).
$$
As this holds for $(D'',\alpha'')$ in a neighbourhood of $(D^2,\alpha^2)$, $(D^2,\alpha^2)$ is a local maximizer of $Q_2$.

\noindent$(\Longleftarrow)$ The converse is similar.  $\Box$

\bigskip

From hereon, we assume $\lambda_t=\lambda$, $t=1,...,T+1$.
Before proving Lemma \ref{lem:ME}, we recall the Karush-Kuhn-Tucker (KKT) conditions from convex analysis, here taken from \cite{rockafellar1970}.

\begin{definition}
	Let $C\neq\varnothing$ be a convex subset of $\real^n$. Let $f_i:C\to\real$ be convex functions on $C$ for $0\leq i\leq r$ and affine functions on $C$ for $r+1\leq i\leq m$.  Consider the following problem,
	\[{\rm (P)}\begin{cases}
		 \textnormal{minimize } &f_0(x)\\
		 \textnormal{subject to } &f_i(x)\leq0\quad(1\leq i\leq r)\\
		 &f_i(x)=0\quad(r+1\leq i\leq m)
	\end{cases}\]
	We call {\rm (P)} an {\rm ordinary convex program}.
\end{definition}

\begin{lemma}[KKT Conditions \cite{rockafellar1970}]
	Let {\rm (P)} be an ordinary convex program. Let $\mu\in\real^m,$ $x\in\real^n$. In order for $\mu$ to be a {\rm KKT} vector for {\rm (P)} and $x$ an optimal solution to {\rm (P)}, it is necessary and sufficient that $(\mu,x)$ be a saddle-point of the Lagrangian of {\rm (P)}. Moreover, this condition holds if and only if $x$ and the components $\mu_i$ of $\mu$ satisfy
	\begin{enumerate}[{\rm (i)}]
		\item $\mu_i\geq0$, $f_i(x)\leq0$, and $\mu_if_i(x)=0$ $(1\leq i\leq r)$;
		\item $f_i(x)=0$ $(r+1\leq i\leq m)$;
		\item $0\in[\partial f_0(x)+\sum_{i=1}^m\mu_i\partial f_i(x)]$, the subgradient of the Lagrangian at $x$.
	\end{enumerate}
\end{lemma}

\noindent {\bf Proof of Lemma \ref{lem:ME}.}  Note that the log-likelihood component of the loss function depends on $(D,\alpha)$ through $\varphi(D,\alpha)=\beta$, so that it remains constant over $\varphi^{-1}(\beta).$
Without loss of generality, assume $\alpha>0$, since if a component is negative, we can multiply the variable in question by -1, and if it is zero we can exclude it from the current analysis, as it remains fixed. We further assume $D_\tau>0$ since if it is zero, then again, it is fixed.

We have the following optimization program,
\[{\rm (P_1)}
\begin{cases}
	\textnormal{minimize } &\sum_t||d_t||_q^q+||\alpha||_q^q\\
	\textnormal{subject to } &-d_\tau<0\\
	&(\Pi_{t}d_{L^t})\alpha_{L_j}=\beta_{L_j}\quad(\forall\,\tau,L_j).
\end{cases}
\]
Let $x=(x_\tau)=(\ln d_\tau)$,  $y=(y_{L_j})=(\ln\alpha_{L_j})$, and $B_{L_j}=\ln\beta_{L_j}$. Then ${\rm (P_1)}$ is equivalent to the following ordinary convex program,
\[{\rm (P_2)}
\begin{cases}
	\textnormal{minimize } &f(x,y)=\sum_{\tau}e^{qx_\tau}+\sum_{L_j}e^{qy_{L_j}}\\
	\textnormal{subject to }&f_{L_j}(x,y)=y_{L_j}-B_{L_j}+\sum_tx_{L^t}=0\quad (\forall\,L_j)
\end{cases}
\]
where we have omitted $f_\tau(x,y)=-e^{x_\tau}<0$ $(\forall\tau)$ since it is trivially satisfied.

Consider now the Lagrangian of ${\rm (P_2)}$, $$\Lambda(x,y,(\mu_{L_j}))=f(x,y)+\sum_{L_j}\mu_{L_j} f_{L_j}(x,y).$$ We remark that KKT conditions (i) and (ii) are immediately satisfied, (i) trivially and (ii) by construction. We need only consider (iii).
By the Karush-Kuhn-Tucker theorem, $(x,y)$ is an optimal solution of ${\rm (P_2)}$ if and only if  $\nabla \Lambda(x,y,(\mu_{L_j}))=0$, with gradient with respect to $(x,y)$,
since the subgradient is unique when the function is differentiable.

We have the following derivatives of the Lagrangian $\Lambda$,
\begin{align}
	\frac{\partial\Lambda}{\partial x_{\tau^u}}&=qe^{qx_{\tau^u}}+\sum_{L:L^u=\tau^u}\mu_{L_j}  \label{der:tau} \\ 
	\frac{\partial\Lambda}{\partial y_{L_j}}&=qe^{qy_{L_j}}+\mu_{L_j} .\label{der:L}
\end{align}
$(\Rightarrow)$ Assume $(x,y)$ is a local minimum for ${\rm (P_2)}$. Then the KKT conditions are satisfied, and we can find $\mu$. By (iii) the derivatives (\ref{der:tau}), (\ref{der:L}) are zero, so that we obtain 
\begin{align}
	e^{qx_{\tau^u}}=\sum_{L:L^u=\tau^u}e^{qy_{L_j}}>0.
\end{align}
 A simple calculation yields $d_{\tau^u}^q = \sum_{L:L^t=\tau^u}||\alpha_L||_q^q$.\\
$(\Leftarrow)$ Assume $d_{\tau} ^q= \sum_{L:L^t=\tau}||\alpha_L||_q^q$ for all $\tau$. This defines an optimal solution if we can find $\mu$ such that the KKT conditions are satisfied. But we derived these above.  $\Box.$

\bigskip

\subsection{Proof of Theorem \ref{main:consistency}}
We show that $\mathbb P(\sup_{||u||=c}Q_n(\beta_n^0+\gamma_nu)<Q_n(\beta_n^0))\geq1-\epsilon$,
where $\gamma_n=\sqrt{p_n}(1/\sqrt{n}+\lambda_n)$, $c>0$ constant. Let $w_L(x)=1$ if $\varphi_{L_1}(D(\beta^0 + xu),1))=0$, $\varphi_{L_1}(D(\beta_n^0+xu),1)$ otherwise. Then
\begin{align*}
	\Delta Q &=Q_n(\beta_n^0+\gamma_nu)-Q_n(\beta_n^0)\\
		&= \ell_n(\beta_n^0+\gamma_nu)-\ell(\beta_n^0)-n\lambda_n\sum_L\left(\frac{||\beta_n^0+\gamma_nu||}{w_L(\gamma_n)}-\frac{||\beta_n^0||}{w_L(0)}\right)\\
		&= \Delta\ell_n-n\lambda_n\Delta N.
\end{align*}

By third order Taylor series expansion, 
\begin{align*}
\Delta\ell&=\gamma_nu^t\nabla_{\beta_n}\ell(\beta_n^0)+u^t\nabla_{\beta_n}^2\ell_n(\beta_n^0)u\gamma_n^2/2+u^t\nabla_{\beta_n}(u^t\nabla_{\beta_n}^2(\ell_n\beta_n^*)u)\gamma_n^3/6\\
	&= I_1+I_2+I_3 .
\end{align*}
We look at each term individually.

By regularity condition (A2),
\begin{align*}
	|I_1| &= |\gamma\nabla_{\beta_n}^t\ell(\beta_n^0)u|\\
		&\leq\gamma_n||\nabla_{\beta_n}^t\ell_n(\beta_n^0)||\cdot||u||\\
		&\ll_p\gamma_n\sqrt{np_n}||u|| - \gamma_n(\sqrt{p_n/n}+\sqrt{p_n}\lambda_n/2\sqrt{c_1})n||u||\\
		&\ll_p\gamma^2_nn||u||.
\end{align*}

For the second term, 
\begin{align}
	I_2 &=\frac12u^t\left(\frac1n\nabla_{\beta_n}^2\ell_n(\beta_n^0)+I_n(\beta_n^0)\right)un\gamma_n^2
	+\frac12u^tI_n(\beta_n^0)un\gamma^2.
\end{align}
We find by the Chebyshev inequality,
\begin{align*}
	\mathbb P\left(\left|\left| \frac1n\nabla_{\beta_n}^2\ell_n(\beta_n^0) + I_n(\beta_n^0) \right|\right|\geq\frac\epsilon{p_n}\right) & \leq \frac{p_n^2}{n^2\epsilon^2}\mathbb E\left(\sum_{i=1}^{p_n}\sum_{j=1}^{p_n}\left(\frac{\partial^2\ell_n(\beta_n^0)}{\partial\beta_{ni}\partial\beta_{nj}}-\mathbb E\left(\frac{\partial^2\ell_n(\beta_n^0)}{\partial\beta_{ni}\partial\beta_{nj}}\right)\right)^2\right)\\
	&=\frac{p_n^2}{n^2\epsilon^2}\sum_{i=1}^{p_n}\sum_{j=1}^{p_n}\mathbb E\left(\frac{\partial^2\ell_n(\beta_n^0)}{\partial\beta_{ni}\partial\beta_{nj}}\right)^2 
	- \mathbb E^2\left(\frac{\partial^2\ell(\beta_n^0)}{\partial\beta_{ni}\partial\beta_{nj}}\right)\\
	&< \frac{p_n^2}{n^2\epsilon^2}\sum_{i=1}^{p_n}\sum_{j=1}^{p_n}c_4\\
	&=\frac{p_n^4}{n}\cdot\frac{c_4}{n\epsilon^2} \ll \frac{p_n^4}{n^2}= o(1)
\end{align*}
so that $I_2=-u^tI_n(\beta_n^0)un\gamma_n^2 + o_p(1)$.

For the third term,
\begin{align*}
	I_3 &= \frac16\sum_L\sum_{l=1}^{|L|}\sum_K\sum_{k=1}^{|K|}\sum_J\sum_{j=1}^{|J|}\frac{\partial^3\ell_n(\beta^*)}{\partial\beta_{L_l}\partial\beta_{K_k}\partial\beta_{J_j}} \\
	&\leq\frac16\sum_{i=1}^n\left(\sum_L\sum_{l=1}^{|L|}\sum_K\sum_{k=1}^{|K|}\sum_J\sum_{j=1}^{|J|}M_{nL_lK_kJ_j}(Y_{ni},X_{ni})\right)^{1/2}||u||^3n\gamma_n^2\\
	&\ll_p np_n^{3/2}||u||ç3\gamma_n^3\\
	&\ll_p p_n^{3/2}\sqrt{p_n}(\lambda_n+1/\sqrt{n})||u||^3n\gamma_n^2\\
	&\ll_p p_n^2(\lambda_n+1/\sqrt{n})||u||^3n\gamma_n^2\\
	&= o_p(n\gamma_n^2)||u||^3,
\end{align*}
since $p_n^4/n\to0$ and $p_n^2\lambda_n\to0$ by hypothesis. Here $L,K,J$ run over all lineages.

Thus $\Delta\ell\ll_p \gamma_n^2n||u||-u^t{\mathcal I}_n(\beta_n^0)un\gamma_n^2/2+o_p(n\gamma_n^2)||u||^3$. Choosing $c$ sufficiently large, $I_2$ dominates $I_1$ uniformly on $||u||=c$, then choosing $n$ sufficiently large, $I_2$ dominates $I_3$ uniformly on $||u||=c$.

We next turn to $\Delta N$. We have $\Delta N = \sum_L\Delta N_L$ where $\Delta N_L = \frac{||\beta_{nL}^0+\gamma_nu_L||}{||w_L(\gamma_n)||} - \frac{||\beta_{nL}^0||}{||w_L(0)||}$.

Suppose $\beta_{nL}^0\neq0$. Then 
$$
|\Delta N_L| =\left| \frac{||\beta_{nL}^0+\gamma_nu_L||}{w_L(\gamma_n)}-\frac{||\beta_{nL}^0||}{w_L(0)}\right|
\leq \left|\frac{||\beta_{nL}^0+\gamma_nu_L||-||\beta_{nL}^0||}{\min\{w_L(\gamma_n),w_L(0)\}}\right|.
$$
For $n$ large enough, $w_L(\gamma_n)\geq(1-\xi)w_L(0)$, $\xi>0$ small, so
$$
|\Delta N_L|<\left|\frac{||\beta_{nL}^0+\gamma_nu_L||-||\beta_{nL}^0||}{(1-\xi)w_L(0)}\right|
\leq \frac{\gamma_n||u_L||}{(1-\xi)w_L(0)}\leq\frac{\gamma_n{||u_L||}^{1/(T+1)}}{1-\xi}.
$$
We have 
$$
n\lambda_n\sum_L\frac{\gamma_n{||u_L||^{1/(T+1)}}}{1-\xi}\leq \frac{\sqrt{n}\gamma_n{||u||^{1/(T+1)}}}{1-\xi}
$$
where $L$ runs over all lineages such that $\beta_{nL}^0=0$. From this it follows that $I_2$ dominates $\sum_{L}|\Delta N_L|$.

Suppose now $\beta_{nK}^0=0$ for lineage $K$. Then 
$$
\Delta N_K = \frac{||\beta_{nK}^0+\gamma_nu_K||}{w_K(\gamma_n)}-\frac{||\beta_{nK}^0||}{w_K(0)}=\frac{\gamma_n||u_K||}{w_K(\gamma_n)}>0.
$$
Therefore the term $(I_2-n\lambda_n\sum_K\Delta N_K)<0$ dominates in $\Delta Q_n$, where $K$ runs over all true sparse lineages, $\beta_{nK}^0=0$, and we have convergence in probability.
Therefore $||\hat\beta_n-\beta_n^0||\ll_p\gamma_n $, which completes the proof of consistency. $\Box$

\subsection{Proof of Theorem \ref{main:grpsel}}

We turn now to the sparsity property. Recall that $\beta_{nL_l}$ is a signed polynomial in $|\alpha_{nK_k}|,$ $k=1,...,|K|$, over lineages $K$. Consider the polynomial map $f:\mathbb R^n\to\mathbb R^n$ given by $\alpha_n\mapsto\beta_n$, where $\alpha_n$ is the projection onto the second entry of $\psi(\beta_n)=(D_n,\alpha_n)$. When $\tilde\alpha_n$ (equivalently $\tilde\beta_n$) is non-sparse, the tangent map $\partial_{\tilde\alpha_n}f:T_{\tilde\alpha_n}(\mathbb R^n)\to T_{\tilde\beta_n}(\mathbb R^n)$ is non-singular. Since the entries $\frac{\partial\beta_{K_k}}{\partial\alpha_{L_l}}$ are themselves polynomials, they are continuous and hence $\partial_{\tilde\alpha_n}f$ is continuous with respect to $\alpha_n$ in a neighbourhood of $\tilde\alpha_n$.

We have 
\begin{align*}
\partial_{\alpha_n}f\frac{\partial}{\partial\beta_n}Q_n&=\partial_{\alpha_n}f\frac{\partial\ell_n}{\partial\beta_n}-n\lambda_n\partial_{\alpha_n}f\frac{\partial}{\partial\beta_n}\sum_{L}\frac{||\beta_L||}{w_L}\\
&=\partial_{\alpha_n}f\frac{\partial\ell_n}{\partial\beta_n}-n\lambda_n\frac{\partial}{\partial\alpha_n}||\alpha_n||\\
&=\partial_{\alpha_n}f\frac{\partial\ell_n}{\partial\beta_n}-n\lambda_n\textnormal{ sign}(\alpha_n)
\end{align*}
where sign$(\alpha_n)$ is understood as the vector of signs (sign($\alpha_{n1}$), ... , sign($\alpha_{np_n}$)).

Assume hereon that $||\beta_n^0-\beta_n||\ll_p \sqrt{p_n/n}$.

By second order Taylor expansion, we have 
\begin{align*}
	\frac{\partial \ell_n}{\partial \beta_{L_l}} =& \frac{\partial\ell_n}{\partial\beta_{L_k}}\Big|_{\beta_n^0}
	+\sum_K\sum_{k=1}^{|K|}\frac{\partial^2\ell_n}{\partial\beta_{K_k}\partial\beta_{L_l}}\Big|_{\beta_n^0}(\beta_{nK_k}-\beta_{nK_k}^0) \\
	& +\sum_{J}\sum_{j=1}^{|J|}\sum_{K}\sum_{k=1}^{|H|}\frac{\partial^3\ell_n}{\partial\beta_{K_k}\partial\beta_{J_j}\partial\beta_{L_l}}\Big|_{\beta_n^*}(\beta_{nJ_j}-\beta_{nJ_j}^0)(\beta_{nK_k}-\beta_{nK_k}^0)\\
	&= G_1+G_2+G_3
\end{align*}
where $J,K$ run over all lineages and $\beta_n^*$ lies between $\beta_n^0$ and $\beta_n$.

We have from the proof of consistency that $G_1\ll_p \sqrt{np_n}$. Bounds for $G_2$ and $G_3$ are provided in \cite{fan2004}, which we derive here for completeness. For $G_2$, we have that
\begin{align*}
G_2=&\sum_{J}\sum_{j=1}^{|J|}\Big[\frac{\partial^2\ell_n}{\partial\beta_{n,J_j}\partial\beta_{nK_k}}\Big|_{\hat\beta_n} - \mathbb E\left(\frac{\partial^2\ell_n}{\partial\beta_{n,J_j}\partial\beta_{nK_k}}\Big|_{\hat\beta_n}\right)\Big]\cdot(\hat\beta_{nJ_j}-\beta_{J_j}^0)\\
	&+\sum_{J}\sum_{j=1}^{|J|}\mathbb E\left(\frac{\partial^2\ell_n}{\partial\beta_{n,J_j}\partial\beta_{nK_k}}\Big|_{\hat\beta_n}\right)(\hat\beta_{nJ_j}-\beta_{J_j}^0)\\
	=&F_1+F_2
\end{align*}
By Cauchy-Schwarz inequality and $||\hat\beta_n-\beta_n^0||\ll_p \sqrt{p_n/n}$, 
$$
|F_2|=\Big|n\sum_{J}\sum_{j=1}^{|J|}I_{n,J_jK_k}(\beta_n^0)(\hat\beta_{nJ_j}-\beta_{nJ_j}^0)\Big|
\ll_p n(\sqrt{p_n/n})\left(\sum_{J}\sum_{j=1}^{|J|}I_{nJ_jK_k}(\beta_n^0)\right)^{1/2}.
$$
By the regularity condition (A2) on the eigenvalues of $I_n$, we have $\sum_{J}\sum_{j=1}^{|J|}I_{nJ_jK_k}\ll_p 1$, so that $|F_2|\ll_p \sqrt{np_n}$. For $F_1$, by the Cauchy-Schwarz inequality, we have 
$$
|F_1|\leq||\hat\beta_n-\beta_n^0||\left[\sum_{J}\sum_{j=1}^{|J|}\left(    \frac{\partial^2\ell_n}{\partial\beta_{n,J_j}\partial\beta_{nK_k}}\Big|_{\hat\beta_n} - \mathbb E\left\{\frac{\partial^2\ell_n}{\partial\beta_{n,J_j}\partial\beta_{nK_k}}\Big|_{\hat\beta_n}\right\}  \right)\right]^{1/2}
$$
so that by regularity condition (A2), $|F_1|\ll_p \sqrt{p_n/n}\sqrt{np_n}\ll_p np_n$. Thus $|G_2|\ll_p \sqrt{np_n}$. For $G_3$,
\begin{align*}
	2G_3=&\sum_J\sum_{j=1}^{|J|}\sum_H\sum_{h=1}^{|H|}\frac{\partial^3\ell_n(\beta_n^*)}{\partial\beta_{H_h}\partial\beta_{J_j}\partial\beta_{K_k}}(\hat\beta_{nJ_j}-\beta_{nJ_j}^0)(\hat\beta_{nH_h}-\beta_{nH_h}^0)\\
	=&\sum_J\sum_{j=1}^{|J|}\sum_H\sum_{h=1}^{|H|}\left[\frac{\partial^3\ell_n(\beta_n^*)}{\partial\beta_{H_h}\partial\beta_{J_j}\partial\beta_{K_k}}-\mathbb E\left\{\frac{\partial^3\ell_n(\beta_n^*)}{\partial\beta_{H_h}\partial\beta_{J_j}\partial\beta_{K_k}}\right\} \right](\hat\beta_{nJ_j}-\beta_{nJ_j}^0)(\hat\beta_{nH_h}-\beta_{nH_h}^0)\\
	&+\sum_J\sum_{j=1}^{|J|}\sum_H\sum_{h=1}^{|H|}\mathbb E\left\{\frac{\partial^3\ell_n(\beta_n^*)}{\partial\beta_{H_h}\partial\beta_{J_j}\partial\beta_{K_k}}\right\}(\hat\beta_{nJ_j}-\beta_{nJ_j}^0)(\hat\beta_{nH_h}-\beta_{nH_h}^0)\\
	=&F_3+F_4.
\end{align*}
By Cauchy-Schwarz inequality and the regularity conditions on $\ell_n$, 
\begin{align*}
F_3^2&\leq\sum_J\sum_{j=1}^{|J|}\sum_H\sum_{h=1}^{|H|}\left[\frac{\partial^3\ell_n(\beta_n^*)}{\partial\beta_{H_h}\partial\beta_{J_j}\partial\beta_{K_k}}-\mathbb E\left\{\frac{\partial^3\ell_n(\beta_n^*)}{\partial\beta_{H_h}\partial\beta_{J_j}\partial\beta_{K_k}}\right\} \right]^2||\hat\beta_{n}-\beta_{n}^0||^4\\
&\ll_p(p_n2/n^2)\cdot\mathcal(n^2)=o_p(np_n)
\end{align*}
and
$$
|F_4|\leq C_5^{1/2}np_n||\hat\beta_n-\beta_n^0||^2\ll_p(p_n^2)=o_p(\sqrt{np_n}),
$$
hence $G_3\ll_p \sqrt{np_n}$ and $G_1+G_2+G_3\ll_p \sqrt{np_n}$.

Consider now the elements of the tangent map $D_{\alpha_n}f$, $\frac{\partial\beta_{nL_l}}{\partial\alpha_{nK_k}}$. We may explicitly write out $\beta_{nL_l}$ in terms of $\alpha_n$,
\begin{align}
\beta_{nL_l} = \alpha_{nL_l}\cdot\prod_{t=1}^T\sum_{K:K^t=L^t}||\alpha_K||.\label{eqn:poly}
\end{align} 
From equation (\ref{eqn:poly}), we obtain
\begin{align*}
\frac{\partial\beta_{nL_l}}{\partial\alpha_{nK_k}}=
\begin{cases}
0&\textnormal{if }L^t\neq K^t\,\forall\,t\\
\alpha_{nL_l}\cdot\textnormal{sign}(\alpha_{nL_l})\cdot\sum_{t=1}^T\prod_{u=1,u\neq t}^T\sum_{J:J^u=L^u}||\alpha_{nJ}||\\
\qquad+\textnormal{ sign}(\alpha_{nL_l})\cdot\prod_{t=1}^T\sum_{J:J^t=L^t}||\alpha_{nJ}||&\textnormal{if $L_l=J_j$}\\
\alpha_{nL_l}\cdot\textnormal{sign}(\alpha_{nK_k})\cdot\sum_{t\in S}\prod_{u=1,u\neq t}^T\sum_{J:J^u=L^u}||\alpha_{nJ}||&\textnormal{otherwise, with}\\&S=\{t:L^t=K^t\}.
\end{cases}
\end{align*}
Suppose that $\beta_{n\tau^t}^0=0$ for some taxon $\tau^t$ of taxon level $t$. Then on inspection for $L^t=\tau^t$, we find $\frac{\partial\beta_{nL_l}}{\partial\alpha_{nK_k}}\ll_p  ({p_n/n})^{1/(2T+2)}$.
From $(D_{\alpha_n}f)^{-1}(D_{\alpha_n}f)=I_{p_n}$, we obtain the relationship $\sum_{j=1}^{p_n}\frac{\partial\beta_{ni}}{\partial\alpha_{nj}}\cdot\frac{\partial\alpha_{nj}}{\partial\beta_{ni}}\ll_p (p_n/n)^{1/(2T+2)} \sum_{j=1}^{p_n}\frac{\partial\alpha_{nj}}{\partial\beta_{ni}}$, so that $(n/p_n)^{1/(2T+2)}\ll_p \sum_{j=1}^{p_n}\frac{\partial\alpha_{nj}}{\partial\beta_{ni}}$.

Without loss of generality, assume $\alpha_n>0$, since we can always transfer the sign from the coefficients to the covariates at every $n$. Evaluated at our $\sqrt{n/p_n}$-consistent estimate $\hat\beta_n$, we have 
\begin{align*}
0&=(D_{\alpha_n}f)^{-1}(D_{\alpha_n}f)\nabla_{\beta_n}\ell_n(\hat\beta_n)-n\lambda_n(D_{\alpha_n}f)^{-1}\textnormal{ sign}(\alpha_{n})\\
&=\nabla_{\beta_n}\ell_n(\hat\beta_n)-n\lambda_n(D_{\alpha_n}f)^{-1}\textnormal{ sign}(\alpha_{n}).
\end{align*}
Thus for a sparse taxon $L_l \in \tau$,
\begin{align*}
\nabla_{\beta_n}\ell_n(\hat\beta_n) &= n\lambda_n(D_{\alpha_n}f)^{-1}\textnormal{ sign}(\alpha_{n})\\
n\lambda_n(n/p_n)^{1/(2T+2)} &\ll_p n\lambda_n\mathcal \sum_{j=1}^{p_n}\frac{\partial\alpha_{nj}}{\partial\beta_{ni}}
\ll_p \sqrt{np_n}  \\
\end{align*}
and hence $n \ll_p p_n^{T+2}$, a contradiction. Therefore $\lim_{n\to\infty}\mathbb P(\delta(\hat\beta_{n\tau})=\delta(\beta_{n\tau}^0))=1$ for all taxa $\tau$. $\Box$

\subsection{Proof of Theorem \ref{main:oracle}}

\textit{\rm (i)} The proof of model consistency is immediate from Theorem \ref{main:grpsel}.

\noindent\textit{\rm (ii)} Let $\mathcal A_n\subseteq\{1,...,p_n\}$ be the subset of indices for nonzero parameters. We have that there exists a $\sqrt{n/p_n}$ maximizer $\hat\beta_n=(\hat\beta_{n\mathcal A_n},0)$ of $Q_n$. We take to writing $Q_n(\beta_{n\mathcal A_n})=Q_n(\beta_{n\mathcal A_n},0)$ through abuse of notation. The following argument concerning the log-likelihood $\ell_n$ follows \cite{zhou2010}.

By Taylor expansion of $\nabla_{\beta_n} Q_n$ about $\beta_{n\mathcal A_n}^0$, we find
\begin{align*}
\frac1n\Big(\nabla_{\beta_n}^2&\ell_n(\beta_{n\mathcal A_n}^0)(\hat\beta_{n\mathcal A_n}-\beta_{n\mathcal A_n}^0) - \nabla_{\beta_n} p_\lambda(\hat\beta_{n\mathcal A_n}) \Big)
 \\
&=-\frac1n\Big( \nabla_{\beta_n}\ell_n(\beta_{n\mathcal A_n})+ \frac12(\hat\beta_{n\mathcal A_n}-\beta_{n\mathcal A_n}^0)^T\nabla_{\beta_n}^2(\nabla_{\beta_n}\ell_n(\beta_{n\mathcal A_n}^*))(\hat\beta_{n\mathcal A_n}-\beta_{n\mathcal A_n}^0)     \Big)
\end{align*}
from the relationship $0 = \nabla_{\beta_n} Q_n(\hat\beta_n)$.

By Cauchy-Schwarz inequality, 
\begin{align*}
\Big|\Big| \frac1{2n}(\hat\beta_{n\mathcal A_n}&-\beta_{n\mathcal A_n}^0)^T\nabla_{\beta_n}^2(\nabla_{\beta_n}\ell_n(\beta_{n\mathcal A_n}))(\hat\beta_{n\mathcal A_n}-\beta_{n\mathcal A_n}^0)  \Big|\Big|^2\\
&\leq\frac1{n^2}\sum_{i=1}^n||\hat\beta_{n\mathcal A_n}-\beta_{n\mathcal A_n}^0||^4\sum_{j_1\in\mathcal A_n}\sum_{j_2\in\mathcal A_n}\sum_{j_3\in\mathcal A_n}M_{nj_1j_2j_3}^3(X_{ni},Y_{ni})\\
&\ll_p p_n^5/n^2=o_p(1/n)
\end{align*}

By Lemma 8 in \cite{fan2004}, obtain

$$
\left|\left| \left( \frac1n\nabla_{\beta_n}^2\ell_n(\beta_{n\mathcal A_n}^0)+\mathcal I_n(\beta_{n\mathcal A_n}^0)\right)(\hat\beta_{n\mathcal A_n}-\beta_{n\mathcal A_n}^0) \right|\right|
=o_p(1/\sqrt{n}).
$$

We turn now to the penalty $p_\lambda$. Since $||\beta_n||$ is a polynomial in $\alpha_n$ of degree greater than 1, we have $\frac1n\nabla_{\beta_n}||\alpha_n||=o_p(\frac1n||\nabla_{\beta_n}||\beta_n||||)=o_p(p_n/n)$.

Therefore from the Taylor expansion of $\nabla_{\beta_n} Q$, we have
$$
\mathcal I_n(\beta_{n\mathcal A_n}^0)(\hat\beta_{n\mathcal A_n}-\beta^0_{n\mathcal A_n})+o_p(p_n/n)
=\frac1n\nabla_{\beta_n}\ell_n(\beta_{n\mathcal A_n}^0)+o_p(1/\sqrt{n}).
$$
Following \cite{fan2004},
\begin{align*}
	\sqrt nA_n\mathcal I_n^{-1/2}(\beta_{n\mathcal A_n}^0)&\left(\mathcal I_n(\beta_{n\mathcal A_n}^0)(\hat\beta_{n\mathcal A_n}-\beta_{n\mathcal A_n})+o_p(p_n/n)\right)\\
	&=\sqrt nA_n\mathcal I_n^{-1/2}(\beta_{n\mathcal A_n}^0)\left( \frac1n\nabla_{\beta_n}\ell_n(\beta_{n\mathcal A_n}^0)+o_p(1/\sqrt{n})\right)
\end{align*}

\begin{align*}
	\sqrt nA_n\mathcal I_n^{1/2}(\beta_{n\mathcal A_n}^0)(\hat\beta_{n\mathcal A_n}-\beta_{n\mathcal A_n})&\leadsto\sqrt nA_n\mathcal I_n^{-1/2}(\beta_{n\mathcal A_n}^0)\frac1n\nabla_{\beta_n}\ell_n(\beta_{n\mathcal A_n}^0)\\
	&\leadsto N(0,\Sigma)
\end{align*}
where $A_n$ and $\Sigma$ are as described in the statement of the theorem. $\Box$

\baselineskip = 15pt plus 3pt minus 3pt

\normalem
\bibliographystyle{plain}
\bibliography{PhyLASSO}

\end{document}